\documentclass[11pt,letterpaper]{article}

\usepackage[margin=1in]{geometry}
\usepackage[T1]{fontenc}
\usepackage{lmodern}
\usepackage{amsmath}
\usepackage{amssymb}
\usepackage{graphicx}
\usepackage{booktabs}
\usepackage{xcolor}
\usepackage{caption}
\usepackage{listings}
\usepackage[numbers,sort&compress]{natbib}

\definecolor{linkblue}{rgb}{0.21,0.49,0.74}
\usepackage[pagebackref,breaklinks,colorlinks,allcolors=linkblue]{hyperref}
\usepackage[capitalize]{cleveref}
\crefname{section}{Sec.}{Secs.}
\Crefname{section}{Section}{Sections}
\crefname{table}{Tab.}{Tabs.}
\Crefname{table}{Table}{Tables}

\title{Bayesian Optimization in Sequence-to-Architecture Latent Space for Zero-Shot NAS}

\author{Ond\v{r}ej T\'{y}bl\\
Department of Cybernetics FEE\\
Czech Technical University\\
{\tt\small tyblondr@cvut.cz}
\and
Luk\'{a}\v{s} Neumann\\
Department of Cybernetics FEE\\
Czech Technical University\\
{\tt\small lukas.neumann@cvut.cz}
}

\date{}

\begin{document}
\maketitle
\begin{abstract}
Zero-shot Neural Architecture Search removes the prohibitive cost of traditional NAS, but its search process is typically based on the evolutionary algorithm (EA); lacking an explicit model of the objective, it often resorts to a near-random search through mutation. Bayesian Optimization offers a principled alternative by modeling the objective and aggregating information across iterations, but scales poorly to the high-dimensional, discrete, graph-structured spaces of modern NAS, restricting its use to only small networks. In this paper, we bring Bayesian Optimization to zero-shot NAS for large-scale architectures by learning a latent space via a Variational Autoencoder trained to reconstruct a novel prefix encoding of architectures and propose a proxy scalarization that combines several zero-shot proxies into a single Bayesian Optimization objective. After only 10,000 iterations of the proposed search algorithm (8 hours on a single GPU), our method found a network architecture which under the given model parameter count constraints achieves state-of-the-art results on three separate tasks -- image classification, object detection and semantic segmentation.
\end{abstract}

\section{Introduction}
\label{sec:intro}

Zero-shot (training-free) Neural Architecture Search (NAS) methods have been recently gaining popularity, because they democratize NAS by removing the tremendous computational requirements associated with traditional NAS, and as such they allow to search for novel, more efficient and/or more accurate deep network architectures at a fraction of a cost and energy. 
The key component of every zero-shot NAS method is the search algorithm that determines which new candidate architectures should be evaluated, based on estimated performance of network architectures evaluated in previous steps, ideally steering the search towards more promising candidates as the algorithm progresses. The majority of zero-shot methods~\citep{lee2024az, tybl2025training, li2023zico} rely on the Evolutionary Algorithm (EA), because it can handle the large search space of modern deep network architectures. But EA has no explicit model of the objective and as such the results are often very close to a random search; often the only operation used to traverse the search space is the mutation operation, which essentially is just a random perturbation of the current state. 

\begin{figure}
\centering
    \includegraphics[width=\columnwidth]{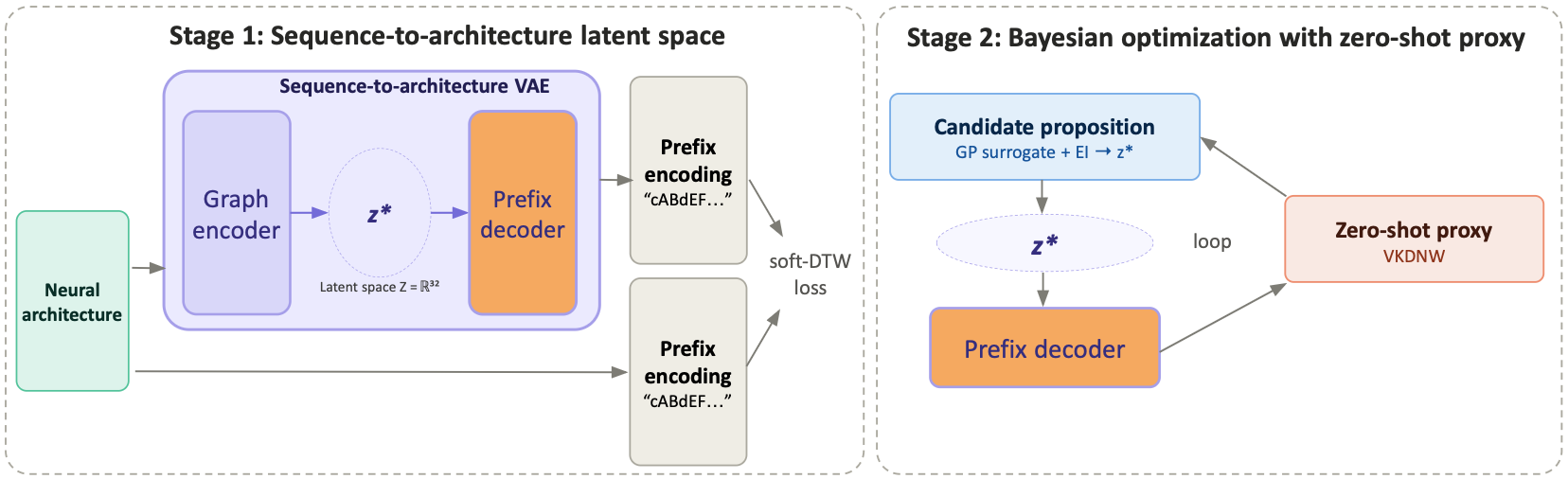}
\caption{We first train a variational autoencoder (VAE) to create a smooth, low-dimensional representation of large-scale heterogeneous architectures (left). Then, the latent space is used for Bayesian Optimization with zero-shot accuracy proxies aggregate as the objective function (right). 
}
\label{fig:figure1}
\end{figure}

Bayesian Optimization (BO), on the other hand is a formal framework that explicitly builds a probabilistic model of the objective function and is therefore able to aggregate information obtained in previous steps to specifically guide the search towards more promising regions of the search space. Bayesian Optimization (BO) however \textbf{scales poorly to high dimensions and discrete, structured spaces}, which unfortunately is exactly the type of search space encountered in NAS methods, where the architectures are discrete graphs consisting of nodes, each of which has several hyper-parameters. This is why the application of Bayesian Optimization in (zero-shot) NAS~\cite{kandasamy2018neural, ru2021interpretable,shen2023proxybo, camero2021bayesian} has so far been limited to only toy problems or small networks, simply because direct application to modern large-scale architectures with hundreds of nodes is computationally prohibitive.

In this paper, we present a novel method that allows to apply Bayesian Optimization (BO) to searching large-scale network architectures in zero-shot NAS, thus eliminating the computational burden of traditional NAS, as well as removing complexity constraints of naively applying BO to the high-dimensional space of large network architectures. We achieve this by introducing a novel latent space that can represent modern large-scale architectures including EfficientNet~\cite{tan2019efficientnet} and Vision Transformer~\cite{dosovitskiy2020image}, by training a Variational Autoencoder (VAE) that learns to reconstruct network architectures expressed by prefix encoding that serializes each architecture block into a compact, bijective string over a vocabulary of primitive and compound operations. The resulting latent space provides a smooth, low-dimensional representation of large-scale heterogeneous architectures suitable as the search domain for Bayesian Optimization.

Additionally, we propose a new proxy scalarization method that allows us to combine several state-of-the-art zero-shot proxies into a single objective. The scalarization whitens and decorrelates the proxy dimensions and, unlike rank-based aggregations, can be evaluated pointwise for any architecture, which is a key prerequisite for modern gradient-free optimization. As a result, our method -- unlike previous methods based on evolutionary algorithms -- is able to learn and exploit the properties of the NAS search space during the search process.

Finally, we use our newly introduced latent space to run Bayesian Optimization, using the scalarized proxy of classification accuracy as the objective function, to search for new architectures. The best architecture as per the proxy value (found without training any of the candidate networks) is only then trained using the standard UniNAS training recipe~\cite{tybl2026universal} to obtain the actual accuracy in three different vision tasks (image classification, object detection/instance segmentation, semantic segmentation). 

In summary, we make the following contributions:
\begin{enumerate}
    \item We introduce a novel latent space that represents modern large-scale architectures including EfficientNet~\cite{tan2019efficientnet} and/or Vision Transformer~\cite{dosovitskiy2020image} simultaneously, which can be used in modern optimization frameworks such as Bayesian Optimization (BO); we build it over UniNAS~\cite{tybl2026universal}, the most challenging neural architecture search space, as it is a superset of the commonly used ones (NAS-Bench 101/201~\cite{ying2019bench,dong2020bench}, Zen-NAS~\cite{lin2021zen}, AutoFormer~\cite{chen2021autoformer}), which are thus special cases already covered by our latent space,
    \item we propose using prefix encoding as the sequential (string-like) representation of network architectures, which yields compact and a bijective mapping to the standard DAG representation. Thanks to it our method is to the best of our knowledge the first latent-space NAS method that works at this scale, as previous ones~\cite{sun2025learn,hemmi2024navigator,hemmi2025graph} rely on graph-based VAEs, which fail to converge in large scale (\Cref{fig:vae_training}),
    \item we propose proxy scalarization method that allows to combine several zero-shot proxies into a single continuous scalar score, so that modern zero-shot approaches can be incorporated into a standard Bayesian Optimization framework.
\end{enumerate}

\vspace{-0.5em}
\section{Related Work}
\label{sec:related_work}

Neural Architecture Search is characterized by two core components: a \emph{performance signal} -- ranging from zero-shot proxies to full training accuracy -- and a \emph{search strategy} that governs how the architecture space is navigated given the signal. We organize prior work along the search-strategy axis, showing how each strategy interacts and scales with the underlying architecture search space.

\paragraph{Evolutionary algorithms (EAs).}
EA leverages population-based optimization to explore architecture spaces without requiring differentiable relaxations. In the classical setting, new candidates are generated by applying mutation operators to the current population. When combined with zero-shot performance proxies -- which bypass full training of candidate architectures -- this paradigm has achieved competitive results~\citep{lee2024az, tybl2025training, li2023zico} on ImageNet-1K~\citep{deng2009imagenet}.
A more recent line of work generates architectures with diffusion: \citet{an2023diffusionnag} steers a graph diffusion model with a performance predictor, and \citet{zhou2025evolution} replaces the learned denoiser with fitness-guided evolution. Both depend on predictors trained on accuracies of many trained architectures, so neither is zero-shot. More broadly, evolutionary methods -- whether mutation- or diffusion-based -- require many architecture evaluations to converge, as they rely on stochastic variation and selection rather than model-guided search. Moreover, because each new candidate is derived from local perturbations of existing ones, the search tends to remain in the neighborhood of the initial population, limiting its ability to discover novel architectures.

\paragraph{Bayesian optimization with hand-crafted encodings.}
Bayesian optimization (BO) offers a principled, sample-efficient alternative to evolutionary search by constructing a surrogate model of the objective function. A central design choice when applying BO to NAS is how architectures are represented. Several works define hand-crafted encodings paired with tailored kernel functions: \citet{kandasamy2018neural} proposed a kernel based on optimal transport over graph representations, evaluated on a space of simple ConvNets; \citet{ru2021interpretable} introduced an interpretable kernel on similar cell-based spaces as a refinement of~\citet{kandasamy2018neural}; \citet{white2021bananas} employed a path-based encoding with an ensemble of neural predictors; and \citet{jin2019autokeras} used an edit-distance kernel over architectures. Zero-shot variants have also been explored, with \citet{shen2023proxybo} applying proxy-based BO on NASBench-201 and \citet{camero2021bayesian} combining BO with a trained random forest surrogate for recurrent networks search. \textit{All of these methods, however, rely on the encoding remaining low-dimensional, which fundamentally limits their scalability} -- none have been demonstrated on large scale datasets such as ImageNet-1K~\citep{deng2009imagenet}. The notable exception is \citet{ouertatani2024efficient}, who operates at ImageNet scale but restricts the search to a narrow set of configuration choices, sidestepping the encoding scalability problem rather than solving it.

Nearly all of these methods have only been validated on small, cell-based search spaces such as NASBench-201; the few ImageNet-1K results rely on transfer from a proxy task and yield modest accuracy~\citep{li2020gpnas, panda2021nastransfer}. More fundamentally, each encoding is tied to a narrow operation set and breaks down when the space grows to 20 or more node types -- as required to jointly cover convolutional and transformer operations. Edit-distance kernels~\citep{jin2019autokeras} face a combinatorial explosion over operation types; path-based encodings~\citep{white2021bananas} assume a small, fixed operation set; vector encodings~\citep{li2020gpnas, ouertatani2024efficient} demand manual redesign for every new vocabulary; and optimal-transport kernels~\citep{kandasamy2018neural} incur costs that grow with both graph size and operation diversity. No existing hand-crafted encoding offers a general, scalable representation for the mixed search spaces that modern architecture design demands.

\emph{Bayesian optimization with learned encodings.}

An alternative to hand-crafting architecture representations is to learn them from data, typically via graph neural networks. Early examples optimize in the latent space of an LSTM sequence encoder--decoder (NAO~\citep{luo2018neural}), a DAG variational autoencoder (D-VAE~\citep{zhang2019d}) or an unsupervised graph autoencoder (arch2vec~\citep{yan2020does}), but only in small search spaces. \citet{hemmi2024navigator,hemmi2025graph} map well-performing architectures into a VGAE latent space and generate candidates for new datasets by randomly perturbing latent points in a PC-DARTS cell space (transferred to ImageNet-1K with accuracy well below $80\%$). \citet{sun2025learn} combine a D-VAE-based graph autoencoder with GP-based BO over NAS-Bench-201 and gradient ascent in its latent space. Graph convolutional networks (GCNs) have also served as encoders: \citet{shi2020bonas} combined BO with a GCN-based predictor, achieving 74.6\% on ImageNet-1K via transfer from NASBench-201, and \citet{ma2019deepgraph} in the context of small images (like CIFAR, MNIST). Existing results thus remain confined to small benchmarks or narrow search spaces, and it is unclear whether learned representations transfer to the large, heterogeneous spaces we target.

\emph{Differentiable architecture search.}
DARTS-based methods~\citep{ye2022b} relax the discrete search space into a continuous one, enabling gradient-based optimization. \citet{ye2022b} achieved 76.1\% on ImageNet-1K. Multi-objective extensions such as NSGANetV2~\citep{lu2020nsganetv2} reaches 80.4\% on ImageNet-1K but requires training a large number of networks to adapt its surrogate. Transfer-based approaches have also been explored: \citet{panda2021nastransfer} applied BO on CIFAR-100 and transferred the discovered architecture to ImageNet, reaching roughly 71\% top-1 accuracy -- underscoring the difficulty of cross-dataset transfer in NAS.

\section{Sequence-to-architecture Latent Space} 
\label{sec:sequence_to_architecture}

In this section, we begin by describing the architecture search space we adapted, followed by our method of encoding network architectures into a unified latent space (\Cref{fig:figure1} -- left). The latent space is then used in Bayesian optimization framework to find network architectures with best accuracy (\Cref{fig:figure1} -- right).
       
\subsection{Architecture space}

We adopt UniNAS~\citep{tybl2026universal} -- a unified generic search space of network architectures -- which contains many well-established hand-crafted topologies including ResNet~\cite{he2016deep}, EfficientNet~\cite{tan2019efficientnet}, ViT~\cite{dosovitskiy2020image}, and CoAtNet~\cite{dai2021coatnet}. At the same time, the search space is also a superset of popular but small-scale NAS benchmarks such as NAS-Bench \cite{dong2020bench,ying2019bench}, Zen-NAS \cite{lin2021zen} and AutoFormer \cite{chen2021autoformer, chen2021searching}, and therefore results achieved in the UniNAS can be viewed as a generalization of the smaller benchmarks.

In UniNAS, each network is represented as a sequence of stages, each containing multiple blocks. Each block is a directed acyclic graph (DAG), which consists of 27 node types that represent elementary operations covering convolution types, matrix multiplications, normalizations, nonlinearities, additions, chunking, and other general tensor operations with varying input/output shapes. The only structural constraint is that the block's input and output must have identical
dimensions and that adjacent nodes in the DAG have matching tensor shapes. 

\subsection{Prefix encoding}
\label{sec:prefix}

The DAG representation of UniNAS nodes is flexible but difficult to encode compactly, and it does not
surface the structured patterns -- such as inverted bottleneck, residual skip connection, multi-head attention -- that characterize modern architectures. We therefore introduce \textit{compound nodes}, which wrap one or more inner sequential operations into a single node (represented as \texttt{Sequential} operation in PyTorch). The child nodes of \textit{compound nodes} are again drawn from the same vocabulary, yielding a recursive grammar. We refer to the number of independent sequential operations of a \textit{compound node} as its \textit{arity}. 

For example, a \textbf{global average pooling node} constitutes a compound node with \textit{arity} of one, because it only wraps a single sequence of operations; a \textbf{branching node} that distributes its input into two branches and then adds the results of these branches is a compound node with \textit{arity} of two, because it has two independent sequential operations. Each compound node type also uniquely determines the spatial dimensionality of children node inputs as well as of the compound node output. Using the same example as above, the children nodes of a \textbf{global average pooling node} always have the input dimension of $\left( C, 1, 1\right)$ due to the spatial dimensionality reduction coming from global average; on the other hand, the children nodes of a \textbf{branch node} have the same input dimension as the parent compound node. For full description of all compound nodes, see \Cref{tab:uninas} in Supplementary Material.

With the introduction of compound nodes, we obtain a more compact version of the UniNAS space with 20 different node types (13 \textit{simple nodes} and 7 \textit{compound nodes}), but at the same time we do not lose the expressive power to encode modern convolution and transformer architectures and their hybrids. In our method, we therefore represent each block as a pair (prefix, shape), where prefix is a string encoding of the block's parse-graph topology described below and shape denotes the block's input (and hence output) dimensions $(C,H,W)$\footnote{Defining only the input shape suffices, since internal node shapes are uniquely determined by propagating through the graph.}.

The encoding exploits the fact that every compound node has a fixed arity, i.e. a predetermined number of inner compound nodes, which makes the serialization self-delimiting without brackets or separators. Specifically, the encoding is defined inductively:

\begin{enumerate}
    \item \textbf{Leaf nodes.} If a DAG consists of a single primitive operation from \Cref{tab:uninas}, it is assigned a unique single character.
    \item \textbf{Sequential composition.} If a DAG is composed of two sub-DAGs in sequence, each having prefix encoding $s_1$ and $s_2$, it is assigned a prefix encoding $s_1 s_2$.
    \item \textbf{Compound nodes.} If a DAG is composed of a compound node having the single-character prefix encoding $c$ with sub-modules with prefix encodings $s_1, \dots, s_n$, then it is assigned a prefix encoding $c s_1 .\cdots s_n.$
\end{enumerate}

As a result, every UniNAS block is associated with a unique encoding (prefix, shape). De-serializing the encoding into an actual computation graph that can be executed on a GPU requires one additional check: that the declared input/output shape can be consistently propagated to all internal nodes. For instance, a block of shape $(4, 16, 16)$ cannot comprise two nested convolutions each reducing the channel count by a factor of $4$, since the intermediate dimensions would become infeasible. This shape-consistency condition is the \emph{only} constraint beyond syntactic well-formedness that must be verified for a prefix string to specify a valid block. This yields a compact, bijective serialization of the full computational DAG topology (see Supplementary Material for worked example).

\subsection{Latent Space Encoding}
\label{sec:latent_variable_encoding}
 
To allow continuous optimization over the space of UniNAS blocks, we train a Variational Autoencoder (VAE) that defines a mapping from latent vectors to block prefix strings. The encoder maps a computational graph $G$ to a diagonal-Gaussian posterior
$q_\phi(\mathbf{z} \mid G) = \mathcal{N}\!\bigl(\boldsymbol{\mu}_\phi(G),\,
\mathrm{diag}\,\boldsymbol{\sigma}^{2}_\phi(G)\bigr)$ in $\mathbb{R}^{d_z}$, where $d_z = 32$;
the decoder autoregressively reconstructs the block's prefix token sequence
$\mathbf{x} = (x_1, \dots, x_T)$ from a sampled latent vector. Both modules are trained end-to-end by maximizing the evidence lower bound 
\begin{equation}\label{eq:elbo}
\begin{split}
    \mathcal{L}(\phi,\theta;\, G)
  \;=&\;
  \mathbb{E}_{q_\phi(\mathbf{z}\mid G)}\!\bigl[\,
    \log p_\theta(\mathbf{x} \mid \mathbf{z})
  \bigr]\\
  &\;-\;
  \beta\,
  D_{\mathrm{KL}}\!\bigl(
    q_\phi(\mathbf{z} \mid G)
    \;\|\;
    \mathcal{N}(\mathbf{0}, \mathbf{I})
  \bigr),
\end{split}
\end{equation} 
where $\beta$ controls the rate--distortion trade-off~\cite{higgins2017betavae}.

\emph{Alignment-aware reconstruction loss.}
Rather than standard position-wise cross-entropy, we use \emph{soft-DTW} loss~\cite{cuturi2017softdtw} as the reconstruction term $\log p_\theta(\mathbf{x}\mid\mathbf{z})$.
Let $\hat{\mathbf{x}} = (\hat{x}_1,\dots,\hat{x}_{T'})$ denote the decoded token sequence and let $e\colon \mathcal{V}\to\mathbb{R}^{d}$ be the shared token embedding.
Soft-DTW first constructs a pairwise cost matrix $\mathbf{C}\in\mathbb{R}^{T'\times T}$ with entries
\begin{equation}\label{eq:sdtw-cost}
  C_{ij} \;=\; \bigl\lVert f_\theta(\hat{x}_i,\,\mathbf{z}) - e(x_j)\bigr\rVert^{2},
\end{equation}
where $f_\theta$ maps each decoded position to the same embedding space.
The loss then computes the minimum-cost monotone alignment between $\hat{\mathbf{x}}$ and $\mathbf{x}$ via the $\gamma$-smoothed soft-minimum recursion (a log-sum-exp relaxation of the standard DTW dynamic programme)~\cite{cuturi2017softdtw}.
 
This alignment-aware formulation confers a concrete advantage over strict position-wise cross-entropy as it is \emph{shift-tolerant}: Suppose that $\hat{\mathbf{x}}$ and $\mathbf{x}$ differ only by a single prefix token inserted at the beginning, that is $\hat{\mathbf{x}}=\left(\hat{x}_1, x_1,\dots,x_{T-1}\right).$
Then, the cross-entropy becomes large as it detects misalignment at each position of the sequence. However, soft-DTW remains small, as this alignment cost is controlled by the edit distance (up to the smoothing parameter $\gamma$) and architecturally minor perturbations remain close in the reconstruction loss~\cite{cuturi2017softdtw,leguen2019dilate}. This is specifically useful in our situation as if we insert a single node into a DAG, its prefix encoding shifts entirely while the corresponding network is actually very similar (one operation more does not usually completely change the architecture).
 
\emph{Encoder.} The encoder treats the block's DAG as an undirected graph -- one node per operation, with bidirectional parent-child edges -- and processes it with $L_{\mathrm{enc}}=3$ graph-convolutional layers of width $d=256$, each followed by a residual connection, GELU activation, and layer normalization.
Global mean-pooling over all node representations yields a single $d$-dimensional vector, which a linear head projects to $\boldsymbol{\mu}\in\mathbb{R}^{d_z}$ and $\log\boldsymbol{\sigma}^{2}\in\mathbb{R}^{d_z}$ for reparameterization sampling.
 
\emph{Decoder.}
The latent vector $\mathbf{z}\in\mathbb{R}^{d_z}$ is expanded into $M=8$ \emph{memory tokens} of dimension $d=256$ by a learned linear projection.
An autoregressive Transformer decoder ($L_{\mathrm{dec}}=4$ layers, $d=256$, $H=4$ heads, FFN width 512) then generates the prefix token sequence one token at a time, attending to the memory tokens and all previously generated tokens via causal masking.
At inference time, decoding is constrained by a finite-state machine (FSM) over the prefix grammar, which guarantees that every emitted string is a syntactically valid prefix without post-hoc repair.
We note, however, that the FSM enforces only syntactic well-formedness of the prefix (i.e.\ conformance to its inductive definition), not semantic feasibility of the resulting DAG: the decoder may still produce prefixes whose corresponding graphs violate shape-matching constraints. Such infeasible samples are discarded during search; in practice, fewer than $5\%$ of decoded prefixes are rejected.
 
\emph{Relation to graph variational autoencoders.}
Because our VAE encodes and decodes a \emph{sequential} prefix string rather than an adjacency matrix, it differs from standard Graph VAEs~\cite{kipf2016variational,simonovsky2018graphvae} in the decoder.
This design choice is deliberate: the sequential representation admits a natural grammar constraint (the FSM above) that is difficult to impose on adjacency-matrix decoders, and it avoids the quadratic scaling of adjacency-based reconstruction with respect to the number of nodes. The encoder, by contrast, aligns with prior graph-VAE encoders.

\vspace{-10pt}
\section{Search}
\label{sec:search}

\subsection{Zero-shot proxy}
\label{sec:zero_shot_proxy}

The principal bottleneck in neural architecture search is the cost of evaluating candidate architectures: na\"ively, each candidate must be trained to convergence before its quality can be assessed.
Zero-shot (aka training-free) NAS methods sidestep this cost by replacing the full training-and-evaluation loop with a lightweight \emph{zero-cost proxy} -- a scalar score computed without training the network, yet demonstrably correlated with the final trained accuracy~\cite{abdelfattah2021zero,lin2021zen,lee2024az,tybl2025training}.

Our method fully embraces this paradigm: the entire search phase optimizes only a zero-cost proxy, with no gradient-based training whatsoever.
Only after the search concludes we train a single architecture -- the top-scoring candidate according to the proxy -- to obtain the final model.
This design makes the search itself orders of magnitude cheaper than approaches that require training each candidate, and concentrates the computational budget on a single, well-chosen training run.

\emph{Training-free proxy.}
We adopt VKDNW~\cite{tybl2025training} as the underlying zero-cost proxy.
VKDNW achieves state-of-the-art rank correlation classification accuracy on multiple benchmarks and search spaces while requiring only a single forward/backward pass through the network on random input data.

VKDNW, like several other modern training-free proxies, is inherently \emph{multi-dimensional}: it produces a vector of scores characterizing different aspects of an architecture (e.g., per-layer or per-block statistics) rather than a single scalar.
To serve as the objective function of a search algorithm, these dimensions must be aggregated into a single scalar that can rank and compare architectures.

A natural baseline is the nonlinear ranking scheme of AZ-NAS~\cite{lee2024az}, which converts each proxy dimension into a rank and multiplicatively penalizes architectures that are not consistently well-ranked across all dimensions.
However, this ranking-based aggregation is designed for search strategies that only require a relative ordering over a fixed, finite population of candidates (e.g., Evolutionary Algorithms operating on the current generation).
Our setting demands a \emph{continuous scalar score} that can be evaluated for any architecture in the search space -- including previously unseen ones -- because our optimization strategy must assess the absolute quality of arbitrary candidates rather than merely rank a pre-enumerated set.
Ranking-based aggregation cannot satisfy this requirement: the rank of a new architecture is undefined without re-ranking the entire population.

\emph{Proxy scalarization.} We therefore propose a Mahalanobis-distance-based scalarization of the multi-dimensional proxy.
Let $\mathbf{p}(\alpha) \in \mathbb{R}^k$ denote the $k$-dimensional vector of $k$ individual training-free proxy values of an architecture $\alpha$, and let $\boldsymbol{\mu}$ and $\boldsymbol{\Sigma}$ denote the empirical mean and whitening matrix of proxy vectors computed over a calibration set of randomly sampled architectures from the search space.
The scalarized proxy score $s(\alpha) \in \mathbb{R}$ is defined as
\begin{equation}
    s(\alpha) = \bigl(\mathbf{p}(\alpha) - \boldsymbol{\mu}\bigr)^\top \boldsymbol{\Sigma}^{-1} 1,
    \label{eq:mahalanobis_scalar}
\end{equation}
where $1$ denotes a vector of all ones, i.e. it is the sum of the standardized (whitened) marginal proxies, closely related to Mahalanobis distance~\cite{mclachlan1999mahalanobis} of the proxy vector from the population mean.

This formulation has two desirable properties.
\emph{(i)~Scale normalization and decorrelation.}\;
Multiplication by $\boldsymbol{\Sigma}^{-1}$ simultaneously standardizes the scale of each proxy dimension---accounting for potentially disparate magnitudes and variances---and decorrelates the dimensions, preventing redundant information shared between correlated proxy components from being double-counted.
Equivalently,~\Cref{eq:mahalanobis_scalar} first applies a whitening transformation~\cite{kessy2018whitening} to the proxy vectors and then sums entries, so that each transformed dimension contributes equally and independently to the final score.
\emph{(ii)~Pointwise evaluability.}\;
Unlike ranking-based methods, our score is a continuous function of the proxy vector and can be evaluated for any architecture without reference to a fixed population.
This makes it compatible with gradient-free optimization, Bayesian optimization, or any other search strategy that evaluates candidates on the fly.

\subsection{Bayesian optimization}
\label{sec:bayesian_optimization}

We search for high-scoring architectures by running Bayesian optimization (BO) directly in the continuous latent space $\mathcal{Z} = \mathbb{R}^{d_z}$, which was obtained by training the VAE (see \Cref{sec:latent_variable_encoding}).
A candidate architecture is produced by decoding a latent vector $z \in \mathcal{Z}$ into a prefix string that is, in turn, decoded into a UniNAS block specification. The block is then used to build a full UniNAS model~\citep{tybl2026universal} following the CoAtNet-style macro-structure~\citep{dai2021coatnet}: a convolutional stem is followed by two stages of increasing channel width and decreasing spatial resolution in which the discovered block template is replicated, followed by two transformer stages. This yields a complete network architecture that is scored with the zero-cost proxies.

We adopt the standard BO framework with a Gaussian process (GP) surrogate~\citep{rasmussen2006gaussian} and expected improvement (EI)~\citep{jones1998efficient,mockus1978application} as the acquisition function; a detailed description of the framework is provided in \Cref{sec:bo_framework} in Supplementary Material.
BO is a well-established tool for hyperparameter tuning~\citep{snoek2012practical} and has more recently been applied to neural architecture search~\citep{kandasamy2018neural,white2021bananas,ru2021interpretable}, but only on small networks due to its computational complexity on larger networks in its naive formulation.

\subsubsection{Design choices for latent-space architecture search}
\label{sec:bo_design}

We now describe the design choices that enable standard BO to operate effectively over the architecture latent space while respecting a target parameter budget.

\emph{Model-size control.}
Because parameter count correlates strongly with proxy performance~\citep{tybl2025training,lee2024az}, controlling for model size is essential: without it, the search would tend to select larger networks rather than architecturally superior ones at a given scale.
We enforce the parameter budget at two complementary levels.

\emph{Soft penalty.}\;
A penalty on deviation from a target parameter count~$n_\star$ is incorporated directly into the objective that enters the GP training set, shaping the surrogate landscape so that it rewards architectures near the desired size.
Concretely, let $s(\alpha)$ denote the scalarized proxy score from~\Cref{eq:mahalanobis_scalar} and $n(z)$ the parameter count of the decoded architecture.
The penalized objective is then
\begin{equation}
\begin{split}
    \tilde{s}(z) \;=&\; s\bigl(\alpha(z)\bigr) \;-\; \pi\!\bigl(n(z)\bigr),\\
    \pi(n)=&\exp\!\Bigl(\frac{\lvert\log(n / n_{\star})\rvert}{\rho}\Bigr) - 1,
\end{split}
    \label{eq:penalized_score}
\end{equation}
where $\rho > 0$ is a penalty-scale hyperparameter.
The exponential form grows super-linearly with the log-ratio deviation from $n_\star$, strongly discouraging large departures while remaining nearly flat in a neighborhood around the target.

\emph{Hard budget mask.}\;
In addition, a hard mask zeros the acquisition function for any candidate whose predicted parameter count falls outside a feasible interval $[n_{\min}, n_{\max}]$, preventing the search from proposing architectures that would exceed memory or compute constraints (see~\Cref{eq:constrained_acquisition} in Supplementary Material).

\emph{Latent decoding with retries.}
Each latent vector $z$ proposed by BO is decoded via the VAE decoder described in \Cref{sec:latent_variable_encoding}.
Because the decoder does not guarantee that every latent point maps to a valid architecture satisfying all shape-matching constraints, we employ a retry mechanism: if the resulting prefix cannot be instantiated as a UniNAS model, we perturb the latent vector up to $M = 100$ times with progressively increasing noise
\begin{equation}
    z^{(t)} = z + \delta\sqrt{t}\varepsilon^{(t)},
    \qquad \varepsilon^{(t)} \sim \mathcal{N}\!\bigl(0,\, I_{d_z}\bigr),
    \label{eq:decode_retry}
\end{equation}
where $\delta>0$ and $t \in \{1, \ldots, M\}$ is the retry index.
The linearly growing variance ensures that early retries remain close to the original proposal (exploiting local smoothness of the latent space) while later retries explore more broadly if the local neighborhood contains no valid architecture.

\section{Experiments}
\label{sec:experiments}

\begin{figure}[h]
    \centering
    \centering
        \includegraphics[width=0.6\linewidth]{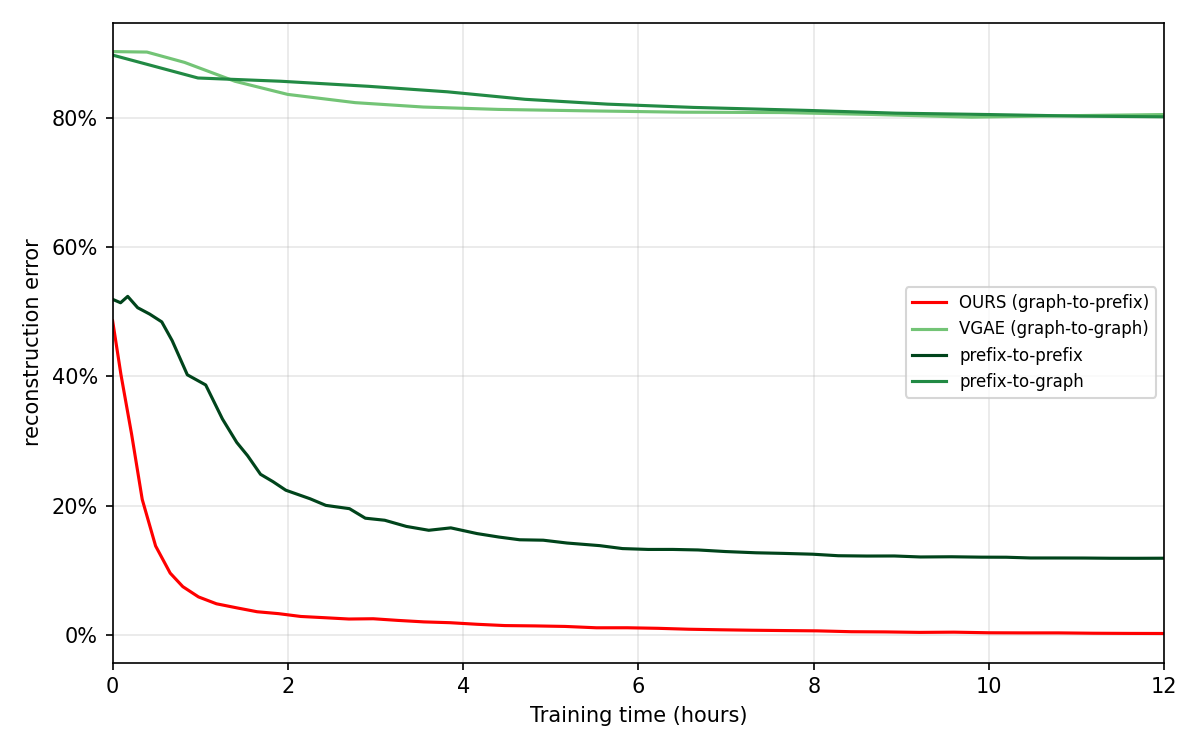}
    \caption{Network architecture reconstruction error (Levenshtein distance) of a variational autoencoder, comparing four encoder–decoder combinations where each component is either graph-based or prefix-sequence-based. The traditinonal VGAE as well as graph-based methods fail to converge, motivating our use of a prefix-sequence decoder.
    }
    \label{fig:vae_training}
\end{figure}

\subsection{Sequence-to-Architecture Latent Space}
\emph{Training VAE.} We train the sequence-to-architecture VAE on ${\sim}96$K randomly generated UniNAS block prefixes, holding out 2\,048 blocks for validation.
Because feasible blocks are generated purely by symbolic graph enumeration, this stage requires no GPU computation.
The VAE uses a maximum token length of~96.
The KL weight is linearly annealed from $\beta = 0$ to $\beta = 10^{-2}$ over the first 5 epochs and held fixed thereafter; training runs for 50 epochs with batch size~128.
We optimize with Adam~\cite{kingma2014adam} using a learning rate of $10^{-3}$, a weight decay of $10^{-4}$, and a gradient clipping at~1.0.
The model has $4.2$M parameters and trains in approximately 12 GPU hours on a single NVidia A100.

\emph{Reconstruction error.}
To evaluate how faithfully the autoencoder preserves architectural structure through the encode--decode round trip, we measure reconstruction quality via the \textit{normalised Levenshtein distance} \citep{levenshtein1966binary} between the decoded and ground-truth token sequences.
Given two token sequences $\mathbf{x}$ and $\hat{\mathbf{x}}$ of lengths $T$ and $T'$, the normalized edit distance is
\begin{equation}
d_\mathrm{lev}(\mathbf{x}, \hat{\mathbf{x}}) = \frac{\mathrm{ed}(\mathbf{x}, \hat{\mathbf{x}})}{\max(T, T')},
\end{equation}
where $\mathrm{ed}(\cdot,\cdot)$ is the Wagner--Fischer edit distance \citep{wagner1974string}, i.e.\ the minimum number of single-token insertions, deletions, and substitutions needed to transform one sequence into the other.
Dividing by the longer sequence length yields a scale-invariant metric bounded in $[0,1]$, where $0$ indicates a perfect token-for-token match.

Our sequence-to-architecture VAE achieves substantially higher reconstruction accuracy than a standard VGAE and comparable graph-based generative baselines, confirming that the token-level formulation better captures the compositional structure of UniNAS blocks (see~\Cref{fig:vae_training}).
This stronger reconstruction is a prerequisite for our approach: because the search operates on prefix-conditioned latent codes, the encoder must faithfully map diverse block topologies into a smooth latent space—a requirement that graph-level embeddings do not satisfy to the same degree.

\subsection{Architecture Search}
\begin{table}
\caption{Top-1 image classification accuracy on ImageNet-1K~\citep{deng2009imagenet}. All models have a similar number of parameters and were trained using the UniNAS training protocol.}
\label{tab:image_classification}
\centering
\footnotesize
\begin{tabular}{lccc}
\hline
\textbf{Model} & \texttt{Params} & \texttt{FLOPs} & \textbf{Top-1} (\%) \\
\hline
ResNet \cite{he2016deep} & 24.8M & 9.0G & 75.19 \\
EfficientNet \cite{tan2019efficientnet} & 24.0M & 6.0G & 80.52 \\
ViT \cite{dosovitskiy2020image, dai2021coatnet} & 26.4M & 18.4G & 80.64 \\
CoAtNet \cite{dai2021coatnet} & 26.9M & 8.44G & 80.58 \\
ZenNAS \cite{lin2021zen} & 26.3M & 18.6G & 80.48 \\
AZ-NAS \cite{lee2024az} & 26.7M & 19.0G & 80.39 \\
ZiCo \cite{li2023zico} & 27.0M & 9.7G & 80.85 \\
UniNAS-A~\cite{tybl2026universal} & 26.8M & 9.0G & 81.15 \\
\textit{ours} & 27.0M & 9.1G & \textbf{81.51} \\
\hline
\end{tabular}
\end{table}

\begin{table*}
\centering
\caption{Comparison of zero-shot (training-free) NAS optimization strategies by best reported top-1 accuracy on ImageNet-1K~\citep{deng2009imagenet}. \dag~denotes methods where optimal architecture was found on a smaller dataset (CIFAR) and then upscaled before training on ImageNet-1K.}
\label{tab:nas_comparison}
\footnotesize
\begin{tabular}{llcc}
\toprule
\textbf{Method} & \textbf{Search strategy} & \textbf{Top-1 (\%)} & \textbf{GPU days} \\
\midrule
$\beta$-DARTS\dag~\cite{ye2022b} & Differentiable (DARTS space) & 76.1 & 0.4 \\
GP-NAS\dag~\cite{li2020gpnas} & Bayesian Opt.\ w/ hand-crafted encoding & 75.8 & 0.9 \\ 
BONAS\dag~\cite{shi2020bonas} & Bayesian Opt.\ w/ learned encoding & 75.4 & 7.5 \\
UniNAS-A~\cite{tybl2026universal} & Evolutionary Algorithm & 81.15 & 0.5 \\
\textit{ours} & Bayesian Opt.\ w/ seq-to-arch latent enc. & \textbf{81.51} & 0.33 \\
\bottomrule
\end{tabular}
\end{table*}

\begin{table*}
\caption{Accuracy of object detection \& instance segmentation on MS-COCO~\cite{lin2014microsoft} and semantic segmentation on ADE20K~\cite{zhou2017scene}. $\text{AP}^{\text{b}}$, $\text{AP}^{\text{m}}$ and $\text{mIoU}$ denote box and mask average precision and mean intersection over union, respectively.}
\label{tab:downstream}
\centering
\footnotesize
\begin{tabular}{lc|cccc|ccc}
\toprule
& & \multicolumn{4}{c|}{\textbf{Object Detection and Seg.}} 
& \multicolumn{3}{c}{\textbf{Semantic Segmentation}} \\
\cmidrule(lr){3-6} \cmidrule(lr){7-9}
Model & \texttt{Params} & AP$^b$ & AP$^m$ & FPS & FLOPs 
& mIoU & FPS & FLOPs \\
\midrule
ResNet \cite{he2016deep}                              & 24.8M & 37.7 & 34.7 & 102.8 & 184G & 39.3 & 481.2 & 47G  \\
EfficientNet \cite{tan2019efficientnet}               & 24.0M & 39.0 & 35.8 & 43.1 & 124G & 37.0 & 152.6 & 32G  \\
CoAtNet \cite{dai2021coatnet}                         & 26.9M & 41.3 & 38.4 & 14.4 & 296G & 42.4 & 61.7 & 51G  \\
UniNAS-A \cite{tybl2026universal}                     & 26.8M & 42.4 & 39.0 & 14.2 & 297G & 45.6 & 88.2 & 51G  \\ 
\textit{ours}                                             & 27.0M & \textbf{42.9} & \textbf{39.8} & 13.5 & 305G & \textbf{46.1} & 62.4 & 51G\\
\bottomrule
\end{tabular}
\end{table*}

\begin{table*}[!t]
  \centering
  \caption{\textbf{Ablation of the main components.} We remove each component and replace it with the closest alternative; the last row is a random-search baseline. All rows use the same search budget as ours (10\,000 evaluated architectures), parameter target and training protocol. $^\dagger$BO did not improve the proxy, and the final architecture training diverged.}
  \label{tab:ablation}
  \small
  \begin{tabular}{lccccc}
    \toprule
    Variant & Search & Latent space & Proxy & Top-1 (\%) & $\Delta$ \\
    \midrule
    Ours (full) & BO & prefix-based & scalarized & \textbf{81.5} & -- \\
    \midrule
    w/o latent space
      & EA & none
      & scalarized & 80.0 & $-1.5$ \\
    w/o proxy scalarization
      & BO & prefix-based
      & VKDNW (single) & 77.2 & $-4.3$ \\
    w/o prefix encoding
      & BO & graph-based (VGAE)
      & scalarized & diverged$^\dagger$ & -- \\
    w/o Bayesian optimization & EA & prefix-based & scalarized & 80.8 & $-0.7$ \\
    baseline & random & none & scalarized & 79.6 & $-1.9$\\
    \bottomrule
  \end{tabular}
\end{table*}

The search is seeded from the union of three complementary architecture pools - randomly grown graphs, random-walk graphs, and mutation-perturbed graphs—designed to provide broad initial coverage of the design space.
We sample $K_{\mathrm{seed}} = 256$ architectures uniformly from this union and evaluate the scalarized zero-shot proxy (see \Cref{sec:zero_shot_proxy}).
The resulting scores are used to fit the scalarizer statistics $\boldsymbol{\mu}$ and $\boldsymbol{\Sigma}$ (\Cref{eq:mahalanobis_scalar}) once; these are frozen for the remainder of the search.
Each seed architecture is encoded to its deterministic posterior mean $\boldsymbol{\mu}_\phi(\mathbf{s})$ and inserted into the BO observation set together with its penalized composite score. We run the process for 10\,000 iterations, resulting in a total 8 GPU hours.

\emph{Image classification.}
After the fixed number of iterations, we collect all evaluated architectures whose parameter count falls within the target budget (centered on the $n_\star$ from~\Cref{eq:penalized_score}, as the search concentrates its evaluations in this region). From this budget-compliant subset, we select the architecture at the $99$th percentile of the penalized objective $\tilde{s}$ rather than the outright maximum, as we find that the very top of the proxy-score distribution is susceptible to noise and proxy miscalibration. The architecture was then trained from scratch on ImageNet-1K~\citep{deng2009imagenet} using the UniNAS training protocol~\cite{tybl2026universal}, which we follow without modification. The training took 32 hours on 8 A100 GPUs.

The model achieves state-of-the-art top-1 accuracy on the ImageNet validation set among models trained with the standard UniNAS protocol (\Cref{tab:image_classification}). Compared to accuracy reported by other zero- and one-shot approaches on the same task (\Cref{tab:nas_comparison}), our method also clearly discovers highest-scoring architecture for the smallest GPU cost, demonstrating that the combination of a well-structured latent space with BO-guided exploration outperforms baselines.

\emph{Downstream tasks.} To assess the generality of the discovered architecture beyond image classification, we fine-tune the ImageNet-pretrained backbone on two downstream tasks -- object detection and instance segmentation on MS-COCO~\citep{lin2014microsoft} and semantic segmentation on ADE20K~\citep{zhou2017scene}. The training took 24 hours, respectively 15 hours on 4 A100 GPUs. All fine-tuning hyper-parameters, data augmentation, and training schedules follow the standard UniNAS protocol~\cite{tybl2026universal}, so that differences in reported metrics can be attributed solely to the choice of backbone architecture.  As demonstrated in \Cref{tab:downstream}, the advantage carries over to downstream tasks: our backbone exceeds competing architectures under the same budget constraints, confirming that the design found by our search generalizes beyond the classification. 

\emph{Ablations.}
In \Cref{tab:ablation}, we replace each component with its closest alternative and add a random-search baseline. All variants use exactly the same budget and settings as ours (10\,000 evaluated architectures, parameter target, selection rule, training protocol), so the differences stem from the replaced component alone.
Without the latent space, BO is not applicable, so we fall back to the evolutionary search (EA) of~\cite{tybl2026universal} in the graph space; accuracy drops to 80.0\% ($-1.5$\,pp).
The same EA in our latent space (w/o Bayesian optimization) reaches 80.8\%, i.e., the latent space alone improves EA by 0.8\,pp and BO adds a further 0.7\,pp, as its surrogate exploits all previous evaluations rather than randomly perturbing the population.
Random search with the same proxy reaches 79.6\% ($-$1.9\,pp), so the gain comes from the guided search, not from the proxy alone.
Without the scalarizer, BO cannot use a composite proxy: unlike EA, its GP surrogate needs an absolute per-architecture score, which rank aggregations do not provide; the strongest single proxy, VKDNW~\cite{tybl2025training}, yields only 77.2\% ($-4.3$\,pp).
Replacing our VAE with the graph-based VGAE of previous latent-space NAS~\cite{hemmi2024navigator,hemmi2025graph} is the most damaging change: its reconstruction error stays at ${\sim}80\%$ (\cref{fig:vae_training}), BO does not improve the proxy, and the selected architecture diverges in training. All components are therefore necessary.

\section{Conclusion}
\label{sec:conclusion}

We proposed a novel zero-shot NAS method which uses Bayesian Optimization in a newly proposed latent space. The key difference from previous latent-space NAS lies in the encoding our autoencoder is trained on: reconstructing the proposed prefix encoding makes it converge on modern large-scale architectures, where graph-decoding autoencoders such as VGAE fail (\cref{fig:vae_training}). The method also allows for using a combination of training-free proxies, thanks to the newly proposed proxy scalarization objective.

We evaluated the method in UniNAS, the most challenging search space, as it is a superset of the commonly used ones: after only 10\,000 iterations of the search (8 hours on a single GPU), it found a network architecture which under the given model parameter count constraints outperforms all compared hand-crafted architectures and NAS methods on three separate tasks -- image classification, object detection/instance segmentation and semantic segmentation; this result was achieved using the official standard UniNAS training recipe, i.e. without any additional bells and whistles or hyper-parameter tuning.

In future work, we plan to evaluate if the latent representations remain effective when paired with alternative search strategies, such as CMA-ES~\cite{hansen2016cma}, beyond the BO framework employed here. Second, our proxy scalarizer is fixed throughout the search, yet the distribution of zero-shot values evolves. Adaptively recalibrating the weights to account for this drift could further improve the search efficiency.

\subsection*{LLM Usage Statement}

In this work, we used Claude by Anthropic (LLM) tools for text polishing, readability improvement, spell-checking and code writing. We did not use generative AI tools for idea generation. We have reviewed all LLM-assisted work and take full responsibility for the final content of this work, including any text, claims, or artifacts produced with the aid of generative LLM.

{
    \small
    \bibliographystyle{plainnat}
    \bibliography{main}
}

\clearpage
\begin{center}
{\Large\bfseries Supplementary Material}
\end{center}

\providecommand{\mainEqPenalty}{Eq.~(4)}
\providecommand{\mainEqRetry}{Eq.~(5)}

\lstdefinestyle{prefixexample}{
  basicstyle=\ttfamily\footnotesize,
  columns=fullflexible,
  keepspaces=true,
  breaklines=false,
  mathescape=true,
  numbers=none,
  frame=none,
}

\section{Graph nodes}
\Cref{tab:uninas} lists the complete node vocabulary used by the prefix encoding of \Cref{sec:prefix}: the 13 leaf nodes retained from UniNAS (the \emph{simple nodes} of the main paper) and the 7 compound nodes.
Four compound nodes encode well-established convolutional patterns: \textsc{ExpandAndReduce}~($f$) applies inner operations at $4\times$ the channel width (inverted bottleneck / transformer FFN); \textsc{ReduceAndExpand}~($m$) does the same at $\frac{1}{4}$ the channels (bottleneck);  \textsc{AvgAndUpsample}~($a$), the \emph{global average pooling node} of \Cref{sec:prefix}, collapses spatial dimensions globally before inner ops then upsamples (the pooled path of squeeze-and-excitation); and \textsc{ForkMerge}(copy,$\times$)~($g$) forks the input to two parallel branches and multiplies their outputs element-wise (gating; with $a$ in one branch this is squeeze-and-excitation, see \Cref{sec:worked_example}). The remaining three compound nodes cover residual, split-and-concatenate and attention structures and thereby open the space to transformer-style and hybrid architectures: \textsc{ForkMerge}(copy,$+$)~($r$), the \emph{branching node} of \Cref{sec:prefix}, adds two branches element-wise, expressing as a single node the residual skip connection that UniNAS builds from separate Copy and Add nodes; \textsc{ForkMerge}(chunk,\,cat)~($o$) channel-splits the input across two branches, each of which therefore receives half of the channels (the branches of the copy-based $g$ and $r$ receive the full input), and concatenates their outputs; and \textsc{ForkMergeAttention}~($h$) encodes multi-head self-attention as a single compound node of arity~5 storing a \emph{head-uniform} per-head template of three branch modules ($Q$/$K$/$V$ projections), one connection module (attention weights), and one post-branch module, replicated $\lfloor C/32\rfloor$ times at model-construction time so that the head count scales automatically with the stage channel width~$C$.  Taken together, the vocabulary can express standard MBConv/EfficientNet blocks, Vision Transformer blocks, and hybrid blocks that freely interleave convolutional and attentional operations---all within a single unified grammar. The grammar describes a single block; how blocks are arranged into stages is fixed by the macro-structure of the main paper and is not part of the encoding.

Copy appears in two rows of \Cref{tab:uninas} because it is folded into both $g$ and $r$; counted once, the 13 retained and the 14 non-retained primitives are the 27 node types of UniNAS.

Of these, only ConvChunk3, Chunk3 and Concat3 have no counterpart in our vocabulary. Dropping these three does not restrict the patterns above: no pattern needs Chunk3 or Concat3, and the one three-way structure among the patterns, the $Q$/$K$/$V$ projections of self-attention, is held by $h$ as three separate branch modules, which makes ConvChunk3 unnecessary as well.

The prefix encoding of \Cref{sec:prefix} is thus defined on this compact version of the UniNAS space: a block is covered if its DAG can be written with the 20 nodes of \Cref{tab:uninas}, i.e.\ if the subsumed primitives occur only inside their compound patterns. UniNAS DAGs outside this set, such as one that merges the two branches of a Copy by Concat2 or one that uses a three-way primitive, have no prefix string. Accordingly, ``every UniNAS block'' in \Cref{sec:prefix} refers to the blocks of this compact space.
\begin{table*}[ht]                                                                                            
\centering
\small                                      
\caption{Node vocabulary and its relation to the original set of UniNAS nodes~\protect\cite{tybl2026universal}.}
\begin{tabular}{@{}clllp{5.2cm}@{}}                                                                          
\toprule                                                                                                     
Letter & Node & Arity & Status & Description \\
\midrule                                                                 
\multicolumn{5}{@{}l}{\textit{Leaf nodes (no children) — 13 retained from UniNAS}} \\[2pt]
$b$ & \textsc{BatchNorm}   & -- & used & batch normalization \\                                                                                                                    
$j$ & \textsc{LayerNorm}   & -- & used & layer normalization \\                                                                                                                    
$c$ & \textsc{Conv1}       & -- & used & $1{\times}1$ convolution \\                                                                                                               
$d$ & \textsc{Conv3}       & -- & used & $3{\times}3$ convolution \\                                                                                                               
$u$ & \textsc{ConvDepth3}  & -- & used & $3{\times}3$ depthwise convolution \\                                                                                                     
$e$ & \textsc{ConvDepth5}  & -- & used & $5{\times}5$ depthwise convolution \\                                                                                                     
$i$ & \textsc{GELU}        & -- & used & GELU activation \\                                                                                                                        
$w$ & \textsc{Sigmoid}     & -- & used & sigmoid activation \\
$v$ & \textsc{MaxPool}     & -- & used & max pooling \\                                                                                                                            
$n$ & \textsc{RelPosBias}  & -- & used & relative positional bias \\
$q$ & \textsc{Softmax}     & -- & used & softmax (attention weights) \\                                                                                                            
$t$ & \textsc{Mask}        & -- & used & attention mask \\                                                                                                                         
$s$ & \textsc{Dropout}     & -- & used & dropout \\
\midrule                                 
\multicolumn{5}{@{}l}{\textit{UniNAS primitives not retained as standalone operations}} \\[2pt]
-- & ConvExp4, ConvRed4        & -- & subsumed & folded into $f$ / $m$ \\  
-- & GlobalAvg, UpSample       & -- & subsumed & folded into $a$ \\       
-- & Copy, Multiply            & -- & subsumed & folded into $g$ \\
-- & Copy, Add                 & -- & subsumed & folded into $r$ \\                                          
-- & Chunk2, Concat2           & -- & subsumed & folded into $o$ \\
-- & Matmul1, Matmul2          & -- & subsumed & folded into $h$ \\
-- & ConvChunk3, Chunk3, Concat3 & -- & discarded & not needed (Q/K/V held by $h$ as separate branches) \\
\midrule
\multicolumn{5}{@{}l}{\textit{Compound nodes (all children are implicit \texttt{Sequential}
                             modules) — all introduced in this work}} \\[2pt]
$f$ & \textsc{ExpandAndReduce}        & 1 & introduced & expand ${\times}4$, inner ops, reduce back \\
$m$ & \textsc{ReduceAndExpand}        & 1 & introduced & reduce ${\div}4$, inner ops, expand back \\
$a$ & \textsc{AvgAndUpsample}         & 1 & introduced & global avg-pool, inner ops, upsample \\
$g$ & \textsc{ForkMerge}(copy,$\times$) & 2 & introduced & two parallel branches; element-wise multiply \\
$r$ & \textsc{ForkMerge}(copy,$+$)   & 2 & introduced & two parallel branches; element-wise add (residual) \\
$o$ & \textsc{ForkMerge}(chunk,\,cat) & 2 & introduced & channel-split; two branches; concatenate \\
$h$ & \textsc{ForkMergeAttention}    & 5 & introduced & per-head template: $Q/K/V$ + connection + post \\                                                             
\bottomrule                                                                 
\end{tabular}                                                                              
\label{tab:uninas}                                                             
\end{table*}

\section{Prefix encoding: worked example}
\label{sec:worked_example}

For illustration, we provide an example of a prefix-encoded DAG following \Cref{sec:prefix}.

We call the sub-modules $s_1,\dots,s_n$ of a compound node in rule~3 (compound nodes) of \Cref{sec:prefix} its \emph{bodies}; they are the implicit \texttt{Sequential} children of \Cref{tab:uninas}, and their number~$n$ is the arity of the node. A body is a sequence of nodes composed by rule~2 (sequential composition) and may hold any number of them, or none, as in an identity branch.

Besides the 20 node letters of \Cref{tab:uninas}, the prefix alphabet contains one further symbol, ``\texttt{.}'', which marks the end of a sequence of nodes. It is not a node type and therefore does not appear in the table: the number of node types stays at 20, while the alphabet has 21 symbols. A compound node with bodies $s_1,\dots,s_n$ is thus written $c\,s_1\texttt{.}\cdots s_n\texttt{.}$, where $c$ is a placeholder for the letter of that node (one of the seven compound letters, not the leaf letter of \textsc{Conv1}). Rule~2 places nothing between the nodes of a sequence, so ``\texttt{.}'' never separates two nodes; it is written only where a whole sequence ends, as a list ends in \emph{nil}. A body without nodes is accordingly the symbol alone, which encodes an identity branch.

With this symbol the prefix string is a Polish (prefix) notation in which every symbol takes a fixed number of sequences as arguments: ``\texttt{.}'' takes none, a leaf letter takes one (the rest of its sequence), and a compound letter of arity~$n$ takes $n+1$ (its bodies and the rest of its sequence). This fixed arity is what makes the serialization self-delimiting without brackets or separators (\Cref{sec:prefix}): the letter of a compound node already tells how many bodies follow, the bodies, each ending in ``\texttt{.}'', simply follow one another, and no bracket pair, separator or explicit count is needed. A string is decoded in one left-to-right pass with a single counter of open sequences, which starts at 1, grows by $n$ at a compound letter of arity~$n$, drops by 1 at ``\texttt{.}'', and reaches zero exactly at the last symbol of a valid string. The block as a whole is itself a sequence, so the complete prefix string of a block is the encoding given by rules~1--3 followed by one final ``\texttt{.}''.

As an example, the following prefix string encodes an inverted-bottleneck block with squeeze-and-excitation gating and a residual connection; its parse tree is given in \Cref{fig:worked_example}.
\begin{lstlisting}[style=prefixexample]
r f e b i g a m i . . w . . . b . . .
\end{lstlisting}

\begin{figure}[t]
\begin{lstlisting}[style=prefixexample]
r                <- ForkMerge(copy,+), arity 2
  [body 1]       <- main branch
  f              <- ExpandAndReduce, arity 1
    [body]       <- expanded (4C) sequence
    e            <- ConvDepth5
    b            <- BatchNorm
    i            <- GELU
    g            <- ForkMerge(copy,$\times$), arity 2
      [body 1]   <- gate
      a          <- AvgAndUpsample, arity 1
        [body]   <- pooled sequence
        m        <- ReduceAndExpand, arity 1
          [body] <- reduced (C) sequence
          i      <- GELU
          .      <- end (body of m)
        .        <- end (body of a)
      w          <- Sigmoid
      .          <- end (body 1 of g)
      [body 2]   <- identity
      .          <- end (body 2 of g, empty)
    .            <- end (body of f)
  b              <- BatchNorm (after reduce)
  .              <- end (body 1 of r)
  [body 2]       <- identity (skip connection)
  .              <- end (body 2 of r, empty)
.                <- end (block)
\end{lstlisting}
\caption{Parse tree of the worked example. The first column, read from top to bottom, is the prefix string. Indentation shows the nesting of bodies; the bracketed labels and the text after ``\texttt{<-}'' are annotations, not symbols of the string.}
\label{fig:worked_example}
\end{figure}

The string thus contains eight ``\texttt{.}'' symbols, one per sequence: the seven bodies (of $m$, $a$ and $f$, and the two each of $g$ and $r$) and the block itself.

\section{Gaussian-process surrogate and acquisition function}
\label{sec:bo_framework}

Bayesian optimization (BO)~\citep{shahriari2016taking,garnett2023bayesian} is a sample-efficient framework for optimizing black-box objectives.
It is built on two components: a \emph{surrogate model} that approximates the objective from a set of evaluations, and an \emph{acquisition function} that decides where to evaluate next.
The surrogate is updated after every new evaluation, and the acquisition function is re-optimized at each step to propose the next candidate.
This loop replaces exhaustive or random search with a principled trade-off between \emph{exploitation}---querying where the surrogate predicts high values---and \emph{exploration}---querying where the surrogate is uncertain.

In our search, the inputs are latent vectors $z \in \mathcal{Z} = \mathbb{R}^{d_z}$ of the sequence-to-architecture VAE, and the value observed at~$z$ is the penalized proxy score $\tilde{s}(z)$ of \mainEqPenalty{} of the main paper, which is maximized.
Throughout this section, $\mu(z)$, $\sigma(z)$ and $k$ denote the GP posterior mean, the GP posterior standard deviation and the kernel, not the like-named quantities of the main paper. Likewise, $f(z)$ and $n(z)$ denote the objective value and the parameter count at~$z$ and are unrelated to the node letters $f$ and $n$ of \Cref{tab:uninas}.

\paragraph{Surrogate: Gaussian process regression.}
The standard surrogate in BO is a Gaussian process (GP)~\citep{rasmussen2006gaussian}.
A GP defines a distribution over functions: given $N$ observations $\mathcal{D} = \{(z_i, y_i)\}_{i=1}^N$, here $y_i$ is $\tilde{s}(z_i)$ after the preprocessing given under \emph{Search settings}, and a zero-mean prior, it provides, at any query point $z$, a posterior mean $\mu(z)$ (the best estimate of the objective) and a posterior variance $\sigma^2(z)$ (the model's uncertainty about that estimate), both available in closed form:
\begin{align}
    \mu(z)      &= \mathbf{k}(z)^\top \bigl(\mathbf{K} + \sigma_\varepsilon^2 I\bigr)^{-1} \mathbf{y}, \label{eq:gp_mean} \\
    \sigma^2(z) &= k(z,z) - \mathbf{k}(z)^\top \bigl(\mathbf{K} + \sigma_\varepsilon^2 I\bigr)^{-1} \mathbf{k}(z), \label{eq:gp_var}
\end{align}
where $\mathbf{K}_{ij} = k(z_i, z_j)$ is the kernel (covariance) matrix over observed inputs, $\mathbf{k}(z)_i = k(z_i, z)$ is the vector of covariances between the observations and the query point, $\mathbf{y} = (y_1, \ldots, y_N)^\top$ collects the observed values, and $\sigma_\varepsilon^2$ is the observation-noise variance.
The kernel $k$ encodes assumptions about the objective's smoothness and scale; its hyperparameters are set by maximizing the log marginal likelihood of $\mathcal{D}$~\citep{rasmussen2006gaussian}.

A widely used kernel family is the Mat\'ern class, parameterized by a smoothness parameter $\nu$, a length-scale $\ell$, and a signal variance $\sigma_f^2$:
\begin{equation}
\begin{split}
    k_{\mathrm{Mat\acute{e}rn}}(z, z';\, \nu, \ell, \sigma_f^2)
    \;=&\;
    \sigma_f^2 \,\frac{2^{1-\nu}}{\Gamma(\nu)}
    \left(\frac{\sqrt{2\nu}\,\|z - z'\|}{\ell}\right)^\nu\\
    &\times\quad K_\nu\!\left(\frac{\sqrt{2\nu}\,\|z - z'\|}{\ell}\right),
\end{split}
    \label{eq:matern_general}
\end{equation}
where $K_\nu$ is the modified Bessel function of the second kind.
The special case $\nu = 5/2$ yields twice-differentiable sample paths and is a standard default in BO, providing a practical middle ground between the very smooth squared-exponential kernel and the rougher $\nu = 3/2$ variant~\citep{snoek2012practical}.
We use this default choice, \cref{eq:matern_general} with $\nu = 5/2$, in our search.

\paragraph{Acquisition function: Expected Improvement.}
The acquisition function translates the GP posterior into a score for each candidate point, encoding the exploitation--exploration trade-off in a single quantity.
At every BO iteration, the next evaluation point is chosen by maximizing this score over the search domain.

We use Expected Improvement (EI)~\citep{jones1998efficient,mockus1978application}, one of the most established acquisition functions.
Let $y^* = \max_i y_i$ denote the best value observed so far (the \emph{incumbent}).
EI is defined as the expected amount by which a new evaluation at $z$ would improve over $y^*$, where the expectation is taken over the GP posterior of the objective value at~$z$, $f(z) \sim \mathcal{N}\!\bigl(\mu(z),\, \sigma^2(z)\bigr)$:
\begin{equation}
    \mathrm{EI}(z)
    \;=\;
    \mathbb{E}\!\Big[\max\!\big(f(z) - y^*,\; 0\big)\Big].
    \label{eq:ei_def}
\end{equation}
The $\max$ ensures that only improvements count; points predicted to be worse than the incumbent contribute zero.
Because the GP posterior at $z$ is Gaussian, this expectation can be evaluated analytically.
Substituting the Gaussian density and integrating yields, for $\sigma(z) > 0$,
\begin{equation}
\begin{split}
    \mathrm{EI}(z)
    \;=\;
    \Delta(z)\,\Phi\!\bigl(u(z)\bigr)
    \;+\;
    \sigma(z)\,\phi\!\bigl(u(z)\bigr),\\
    \Delta(z) = \mu(z) - y^*,
    \qquad
    u(z) = \frac{\Delta(z)}{\sigma(z)},
\end{split}
    \label{eq:ei_closed}
\end{equation}
where $\Phi$ and $\phi$ are the standard Gaussian CDF and PDF.
The two terms have a direct interpretation: $\Delta(z)\,\Phi(u)$ is large when the posterior mean already exceeds the incumbent (\emph{exploitation}), while $\sigma(z)\,\phi(u)$ is large when the posterior uncertainty is high, so that $f(z)$ has a chance of exceeding $y^*$ even if $\mu(z)$ does not (\emph{exploration}).
EI is non-negative everywhere and strictly positive wherever $\sigma(z) > 0$; as $\sigma(z) \to 0$ it tends to $\max\bigl(\Delta(z), 0\bigr)$, so it vanishes only where the model is certain that $z$ does not improve on the incumbent. This makes it a natural criterion for directing the search toward the most informative evaluations.

\paragraph{Constrained acquisition function.}
When the search must respect a hard constraint on a directly
computable quantity (e.g.\ parameter count, FLOPs, or memory
footprint), we augment the acquisition function with a binary
feasibility mask.  In our search this quantity is the parameter count
$n(z)$ of the architecture decoded from~$z$ (the \emph{predicted}
parameter count of the main paper), and the admissible interval is the
budget $[n_{\min}, n_{\max}]$ of the main paper.  The acquisition
function becomes
\begin{equation}
    \mathrm{EI}_{\mathrm{mask}}(z)
    \;=\;
    \begin{cases}
        \mathrm{EI}(z), & n(z) \in [n_{\min},\, n_{\max}],\\
        0,              & \text{otherwise}.
    \end{cases}
    \label{eq:constrained_acquisition}
\end{equation}
The count $n(z)$ is computed exactly, without training, by counting the
parameters of the decoded architecture graph: the mask is a
deterministic function of the decoded architecture and is evaluated
for every proposal that is decoded.  Outside the feasible
interval the acquisition function is exactly zero, which concentrates
the proposals on candidates that satisfy the constraint.  The mask
complements the soft penalty of \mainEqPenalty{} of the main paper,
which acts on the observed values and not on the acquisition
function.

The mask co-exists with the retry mechanism of the main paper.  When
the prefix decoded from a proposal cannot be instantiated and the
proposal is replaced by a perturbed latent $z^{(t)}$ (\mainEqRetry{} of
the main paper), the budget is checked again for the architecture
decoded from $z^{(t)}$.  The interval $[n_{\min}, n_{\max}]$ is thus
enforced on every architecture that BO proposes and scores, whether it
is decoded from the proposal itself or from one of its retries.

\paragraph{Search settings.}
We use $\rho = 0.2$, $n_\star = 27\mathrm{M}$, $[n_{\min}, n_{\max}] = [23\mathrm{M}, 28\mathrm{M}]$ and $\delta = 0.1$. The scalarization combines the five
components of VKDNW (the proxy dimension $k = 5$ of the main paper). The GP is configured with a constant-scaled isotropic
Mat\'ern-$5/2$ kernel plus white noise (initialized at $\sigma_f^2 = 1$,
$\sigma_\varepsilon^2 = 10^{-2}$), on standardized latents and standardized scores
winsorized at the 5th percentile. Its hyperparameters maximize the log marginal likelihood (5 optimizer restarts). During the search, the refits of the GP are approximate rather than exact.

\end{document}